\documentclass[a4paper, 10pt, conference]{ieeeconf}

\usepackage[utf8]{inputenc}
\usepackage[T1]{fontenc}
\usepackage{graphicx,caption,subfig,url,booktabs,gensymb}

\usepackage{xcolor}
\usepackage[printwatermark]{xwatermark}
\newwatermark*[allpages,color=red,angle=0,scale=1,xpos=0,ypos=390pt,fontsize=6pt]{
The content of this paper was published at FG, 2017. This ArXiv version is from before the peer review. Please, cite the following paper:
\\Ziga Emersic, Dejan Stepec, Vitomir Struc, Peter Peer: "Training Convolutional Neural Networks with Limited Training Data for Ear Recognition in the Wild", \\Conference on Automatic Face and Gesture Recognition, -- International Workshop on Biometrics in the Wild IEEE, 2017
}

\graphicspath{{./figures/}}
\DeclareGraphicsExtensions{.pdf,.png,.jpg}

\newcommand\blfootnote[1]{%
  \begingroup
  \renewcommand\thefootnote{}\footnote{#1}%
  \addtocounter{footnote}{-1}%
  \endgroup
}

\usepackage[hang,flushmargin]{footmisc}
\title{\LARGE \bf
Training Convolutional Neural Networks with Limited Training Data for Ear Recognition in the Wild
}

\author{\parbox{16cm}{\centering
    {\large \v{Z}iga Emer\v{s}i\v{c}$^1$, Dejan \v{S}tepec$^1$, Vitomir \v{S}truc$^2$ and Peter Peer$^1$}\\
    {\normalsize
    $^1$ Faculty of Computer and Information Science, University of Ljubljana, Slovenia\\
    Email: \{ziga.emersic, dejan.stepec, peter.peer\}@fri.uni-lj.si\\
    $^2$ Faculty of Electrical Engineering, University of Ljubljana, Slovenia\\
    Email: vitomir.struc@fe.uni-lj.si\\
    }}
}

\begin{document}

\thispagestyle{empty}
\pagestyle{empty}
\maketitle

%%%%%%%%%%%%%%%%%%%%%%%%%%%%%%%%%%%%%%%%%%%%%%%%%%%%%%%%%%%%%%%%%%%%%%%%%%%%%%%%
\begin{abstract}
Identity recognition from ear images is an active field of research within the biometric community. The ability to capture ear images from a distance and in a covert manner makes ear recognition technology an appealing choice for surveillance and security applications as well as related application domains. In contrast to other biometric modalities, where large datasets captured in uncontrolled settings are readily available, datasets of ear images are still limited in size and mostly of laboratory-like quality. As a consequence, ear recognition technology has not  benefited yet from advances in deep learning and convolutional neural networks (CNNs) and is still lacking behind other modalities that experienced significant performance gains owing to deep recognition technology. In this paper we address this problem and aim at building a CNN-based ear recognition model. We explore different strategies towards model training with limited amounts of training data and show that by selecting an appropriate model architecture, using aggressive data augmentation and selective learning on existing (pre-trained) models, we are able to learn an effective CNN-based model using a little more than  1300 training images. The result of our work is the first CNN-based approach to ear recognition that is also made publicly available to the research community. With our model we are able to improve on the rank one recognition rate of the previous state-of-the-art by more than 25\% on a challenging dataset of ear images captured from the web (a.k.a. in the wild).
\end{abstract}

%%%%%%%%%%%%%%%%%%%%%%%%%%%%%%%%%%%%%%%%%%%%%%%%%%%%%%%%%%%%%%%%%%%%%%%%%%%%%%%%%
\section{Introduction}

\blfootnote{\textbf{The content of this paper was presented at the International Workshop on Biometrics in the Wild 2017 as a part of the International Conference on Automatic Face and Gesture Recognition 2017.}\hfill}

Convolutional neural networks (CNNs) have recently demonstrated impressive performance on various computer vision tasks such as semantic image segmentation~\cite{chen2014semantic}, object detection and recognition~\cite{donahue2014decaf,redmon2016you,erhan2014scalable} , image super-resolution~\cite{dong2014learning, cui2014deep, dong2016image} and alike. One of the key factors contributing to this development is the availability of extensive corpora of training data that allow for the design and especially training of deep CNNs. For problems, where training data is in abundance, CNNs have pushed the state-of-the-art to new unprecedented heights, whereas in other areas, where training data is scarce, the impact of deep learning and CNNs has been limited. One such area is automatic person recognition from ear images, which has immense potential in forensic, security and surveillance applications, but lacks the necessary large-scale datasets to make full use of recent developments in deep learning and convolutional neural networks. 

While ear recognition is getting more and more popular over recent years, the existing datasets in this area are still limited to a few thousand images with a few hundred identities and are typically of laboratory-like quality with constrained variability. This is in stark contrast to the field of face recognition (or image classification), where datasets used for CNN training are nowadays measured in tens of millions of images with tens of thousands of identities. Moreover, these datasets are commonly harvested from the web and are therefore generally considered to be representative of real-world settings.   

In this paper we address the problem of training CNNs with limited training data and strive to develop an effective CNN-based model for ear recognition. Existing approaches to CNN training with small amounts of training data typically include \textit{i)} metric-learning approaches, where training is performed with image pairs (or even triplets) instead of single images~\cite{hu2014discriminative,hoffer2015deep}, \textit{ii)} data augmentation techniques that in addition to geometric and color perturbations of the existing training data also include the generation of synthetic data samples~\cite{guo2016deep,guo2015deep, dellana2016data,hu2016frankenstein}, and \textit{iii)} using existing CNNs (trained for related recognition problems) as so-called ``black-box'' feature extractors, on top of which additional classifiers are trained and used for recognition~\cite{grm_deep2017}. Here, we build on these approaches and successfully develop a CNN model for ear recognition by exploring different strategies to network training, i.e.:
\begin{itemize}
\item Model architecture selection: we investigate different CNN architectures with different parameter-space sizes. Specifically, we consider an AlexNet-like architecture~\cite{caffe,alexNet}, the 16-layer VGG model architecture~\cite{vgg}, and the more recent SqueezeNet architecture~\cite{SqueezeNet}. 
\item Aggressive data augmentation: we examine various data augmentation techniques and try to train CNN models for ear recognition from scratch using a (web-harvested) dataset of unconstrained ear images. 
\item Selective model learning: we consider pre-trained CNN trained initially for face or image classification and then learn only parts of the model to reduce the number of parameters that need to be estimated with the available training data.
\end{itemize}
The result of our development work is a CNN-based model for ear recognition that is able to increase the Rank-1 recognition rate on our test data by more than 25\% compared to the existing state-of-the-art. %We make the model publicly available to the research community through: {\small \url{http://www.anonymized_url.com}}. %TODO!!!

In summary, we make the following contributions in this paper:
\begin{itemize}
\item  We develop the first CNN-based model for ear recognition using a limited amount of training data and significantly improve on the existing state-of-the-art on a challenging dataset of ear images gathered from the web.
\item We evaluate, compare and discuss different strategies to training CNN-based recognition models with limited training data and elaborate on what works and what does not.
\item We present comparative experiments between our CNN-based model and 7 state-of-the-art descriptor-based techniques for ear recognition.
\end{itemize}

The rest of the paper is structured as follows. In Section~\ref{section:related_work} we briefly review the existing work on ear recognition. In Section~\ref{section:methodology} we present the background and motivation for our work, describe the CNN architectures and strategies we followed during model learning. In Section~\ref{section:experiments} we outline the experimental setup, datasets, and report experimental results. We conclude the paper in Section~\ref{section:conclusion}, respectively.

%%%%%%%%%%%%%%%%%%%%%%%%%%%%%%%%%%%%%%%%%%%%%%%%%%%%%%%%%%%%%%%%%%%%%%%%%%%%%%%%%
\section{Related work}
\label{section:related_work}

The goal of this section is to provide the reader with a brief high-level overview of existing techniques in the field of ear recognition and CNN-based recognition models. For a comprehensive overview of the two areas, the reader is referred to in-depth surveys on these topics~\cite{guo2016deep,awe}. 

{\bf Ear recognition:} According to Emer\v{s}i\v{c} et al.~\cite{awe} existing techniques for ear recognition can conveniently be grouped into \textit{i)} geometric, \textit{ii)} holistic, \textit{iii)} local, and \textit{iv)} hybrid techniques.% OD tu naprej je treba stavke mal obračat

Geometric techniques describe the geometric properties of ears or derive geometry-related statistics that can be used for recognition. Since only information related to the ear geometry is used, it is straightforward to devise methods that are invariant to geometric distortions, such as rotation, scaling or even small perspective changes. Examples of techniques from this group were presented in~\cite{awe,anika}. 

Holistic techniques rely on the global ear appearance and exploit representations that encode the ear structure as whole. As the appearance of ears varies significantly with pose or illumination, care needs to be taken before computing holistic features from input images and normalization techniques need to be applied to correct for these changes prior to feature extraction. Examples of global techniques can be found in~\cite{awe31,yuan2006ear,zhang2005novel}.

Local approaches extract features from spatially-confined areas of an image without leaning on the global information describing the overall structure of the ears. The extracted features do not necessarily correspond to structurally meaningful parts of the ear, but can in general represent any point in the image.
We distinguish two types of local techniques: techniques that first detect keypoint locations in the image and then compute descriptors for each of the detected keypoints~\cite{sift} and techniques that compute local descriptors densely over the entire image with no regard to the image's structural characteristics. Examples of local techniques include~\cite{awe,guo2008ear,ojansivu2008rotation,vu2010face}.

The last groups of techniques, so-called hybrid techniques,  combine elements from other categories or use multiple representations to increase the ear recognition performance~\cite{awe}. Techniques from this group offer superior performance compared to competing techniques, but often at the cost of higher computational complexity~\cite{awe50,awe75,awe81}. As suggested by recent surveys on ear recognition~\cite{awe,anika,abaza2013survey}, hybrid techniques together  with local descriptor-based methods represent the current state-of-the-art in this area.

{\bf CNN-based recognition models:} 
Various CNN architectures have been developed and presented in the literature over recent years. One of the most well-known problems that highlighted the potential and power of CNN-based approaches was object classification within the ImageNet Large Scale Visual Recognition Challenge (ILSVRC)~\cite{imagenet}. The AlexNet architecture introduced by Krizhevsky et al.~\cite{alexNet} achieved an unprecedented performance on the ImageNet data and triggered a surge in the use and popularity of CNN-based models in computer vision~\cite{alexNet}. The next milestone in terms of ILSVRC results marked the introduction of the 16-layer  VGG~\cite{vgg} architecture which provided an additional boost to the recognition rates of ILSVRC. These two architectures are nowadays also used in other tasks for example as part of  larger architectures such as Faster-RCNNs~\cite{faster_rcnn} or SegNet~\cite{segNet1,segNet2}. A more recent architecture, called ResNet, was presented by He et al.~\cite{he2016deep}. The architecture introduced shortcut connections to CNNs and made it possible to reliable train deep networks with several hundreds or even thousands of network layers. 

To the best of our knowledge, no CNN-based methods for ear recognition have yet been presented in the literature, with the exception of Galdamez et al.~\cite{galdamez2016brief}, where the authors tried to build separate CNN models for each subject. The main reason for the lack of work in the field of CNN-based ear recognition is in our opinion the absence of large-scale datasets and the difficulty of effective training of deep model with small amounts of data as also suggested in~\cite{awe}. 

%%%%%%%%%%%%%%%%%%%%%%%%%%%%%%%%%%%%%%%%%%%%%%%%%%%%%%%%%%%%%%%%%%%%%%%%%%%%%%%%%
\section{Methodology}
\label{section:methodology}

In this work, we are interested in CNN-based models applicable for the task of closed-set identification from ear images. Our goal is to train CNN-based models that are able of determine the correct class (or identity) of an input ear image from a closed-set of predefined target identities. Thus, each  CNN model considered  features a softmax layer that returns a class-membership distribution over all available target classes when an input ear images is fed to the model. In the remainder of this section we outline the main strategies we explore to train an effective CNN-based model for ear recognition based on an limited amount of training data.  %      output predictions to which of a predsubject each test sample belongs to. Furthermore, it means that we used soft-max layers on all three architectures as the last, output layer with the number of outputs the same as the number of classes in our dataset.

\subsection{Learning strategies} %%%%%%%%%%%%%%%%%%%%%%%%%%%%%%%%%%%%%%%%%%%%%%%%%%%%%%%%%%%%%%%%%%%%%%%%%%%%%%%%%

To train a CNN-based model for ear recognition we investigate the following learning strategies.  

%\subsubsection{Model architecture selection} 

\textbf{Model architecture selection:} 
Different CNN-based models contain different numbers of model-parameters that need to be determined during training. This fact imposes certain requirements on the amount of training data that need to be available for parameter learning. While it is reasonably to assume that lighter architectures (i.e., architectures with less parameters) require less data to be trained, the convergence of the back-propagation (learning) procedure also needs to be considered, since certain architectures may facilitate faster learning. To study the impact of the model architecture on the learning procedure, we consider three popular CNN configurations, i.e.,
\begin{itemize}
\item \textit{AlexNet}~\cite{alexNet}, which is one of the first successful CNN architectures initially applied for the problem of object classification on the ImageNet dataset~\cite{imagenet}. AlexNet helped to popularize the field of deep learning and still represents a successful architecture used for various vision tasks. The model contains around 58.96 million parameters and 500,000 neurons and consists of five convolutional layers, (some of which are followed by max-pooling layers), and three fully-connected layers. The fully connected layers are followed by a softmax classifier that outputs class-membership probabilities for each relevant class. In this work, we used an AlexNet-like architecture available with the Caffe library (also referred to as CaffeNet~\cite{caffe}), where additional normalization layers are used after the pooling operations. %For the last, soft-max layer, we used 166 outputs. We used CaffeNet model which is a replication of the model described in the AlexNet publication with some minor differences: not using data augmentation inside the network and pooling is done before normalization. However, in the continuation of the paper we refer to this architecture as AlexNet.
\item \textit{VGG-16}~\cite{vgg} is a representative of so-called \textit{very deep} CNN models. The main characteristic of the network is the use of several consequtive convolutional layers with small $3\times3$ kernels (or filters) -- which is also the smallest filter size capable of encoding directional information. These stacks of $3\times3$ convolutional layers are able to capture the same information as the larger filter used with AlexNet, but require significantly less parameters that need to be estimated during training. The $3\times3$ filter stacks are interspersed with max-pooling layers which reduce the dimensionality of the activation maps produced by the model layers. The convolutional part of the VGG model is followed by three fully-connected layers with 4,096, 4,096 and 1,000 channels respectively. Finally a soft-max layer is used at the top of the network. The VGG model contains around 134.94 million parameters. %before the output and was set to 166 to match our data.%VEČ MALIH FILTROV 3X3, KI APROKSIMIRAJO VELIKO V ZAPOREDJU, VSEENO MANJ PARAMETROV, VEČ KONVOLUCIJ ZAPOREDOMA, POTEM POOLING, PARAMETER SPACE MANJŠI, PODVAJANJE FILTROV PO POOLINGU, OPIS SPECIFICNE KONFIURACIJE, KI SMO JO UPORABILI - GLEJ HAZIMOV CLANEK! ARXIVE OD HAZIMA CVPRW BIOMETRICS 16)~\cite{hazim}.
\item \textit{SqueezeNet}~\cite{SqueezeNet} represents a special example of a lightweight residual network (or ResNets), where squeeze layers, i.e., layers consisting only of $1\times 1$ convolutions, are added to the network with the goal of reducing the model size and number of weights that need to be tuned during training. Furthermore, no fully-connected layers are present in the network, instead an average-pooling layer is used at the top of the network to produce the final representation that serves as the input to the softmax classifier. The SqueezeNet architecture also contains residual connections whose purpose is to make the back-propagation-based learning more efficient. The architecture used for our experiments contains the following layers from input to output: a convolutional layer, a max-pooling layer, three so-called fire modules, a max-pooling layer, another four fire modules, a max-pooling layer, one additional fire module, a convolutional layer and finally the average-pooling layer. In total, SqueezNet model used in our experiments contains around 821 thousand parameters before pruning.
\end{itemize}

\textbf{Full model learning:} Learning CNN-based models from scratch is a difficult task that requires large amounts of appropriately annotated training data. The model parameters are learned with some form of back-propagation algorithm that tries to minimize a loss function defined over the output of the model. Commonly, the lower network layers learn to respond to primitive visual features, such as edges or corners, which are then gradually combined into more complex structures in the higher model layers until some data representation is learned in the fully connected layers of the CNN-based model driven by the learning criterion. 

Since most CNN architecture contain several millions of parameters that need to be learned during training, even large datasets featuring tens of thousands of images are usually not sufficient to facilitate successful training. Aggressive data augmentation is, therefore, a must with CNN-based models. When trying to learn a CNN-based model from scratch, researchers typically augment the available training data by producing data variations with, e.g., geometric transformations, color modifications, addition of noise, and more recently also by synthesizing samples of artificial identities, as, for example, described in~\cite{hu2016frankenstein}.

When investigating the use of \textit{full model learning} for the problem of ear recognition, we also rely on aggressive data augmentation and increase the amount of available training data by up to a factor of 100. Since data augmentation is also used with the selective training (described next), we present our data augmentation steps in a separate section below.

%Learning CNN-based models from scratch .
%In this strategy we learned all the model layers from scratch. This means that we did not take ImageNet's pre-learned weights. The first approach of learning all the layers from scratch is, as far as the strategy goes, straightforward and is the same for all the three architectures. 
%However, for the second case strategies vary in certain aspects and are described below.%ja tole je masilo

\textbf{Selective model learning:} CNN-based models have been successfully applied to a number of visual recognition problems as outlined in the introductory section. These models typically share some common characteristics, which are reflected in the CNN-based models, especially at the lower model layers. It is, therefore, reasonable to assume that CNN-based models trained, for example, for object recognition should also be able to produce useful representations of more specific object classes, such as ears, along the model layers. While these representations are likely not optimal for the problem of ear recognition, existing pretrained CNN-based models should nevertheless represent a suitable initial model configuration that may be exploited to reduce the need for large amounts of training data.
\begin{figure*}
	\center
	\includegraphics[width=1\textwidth]{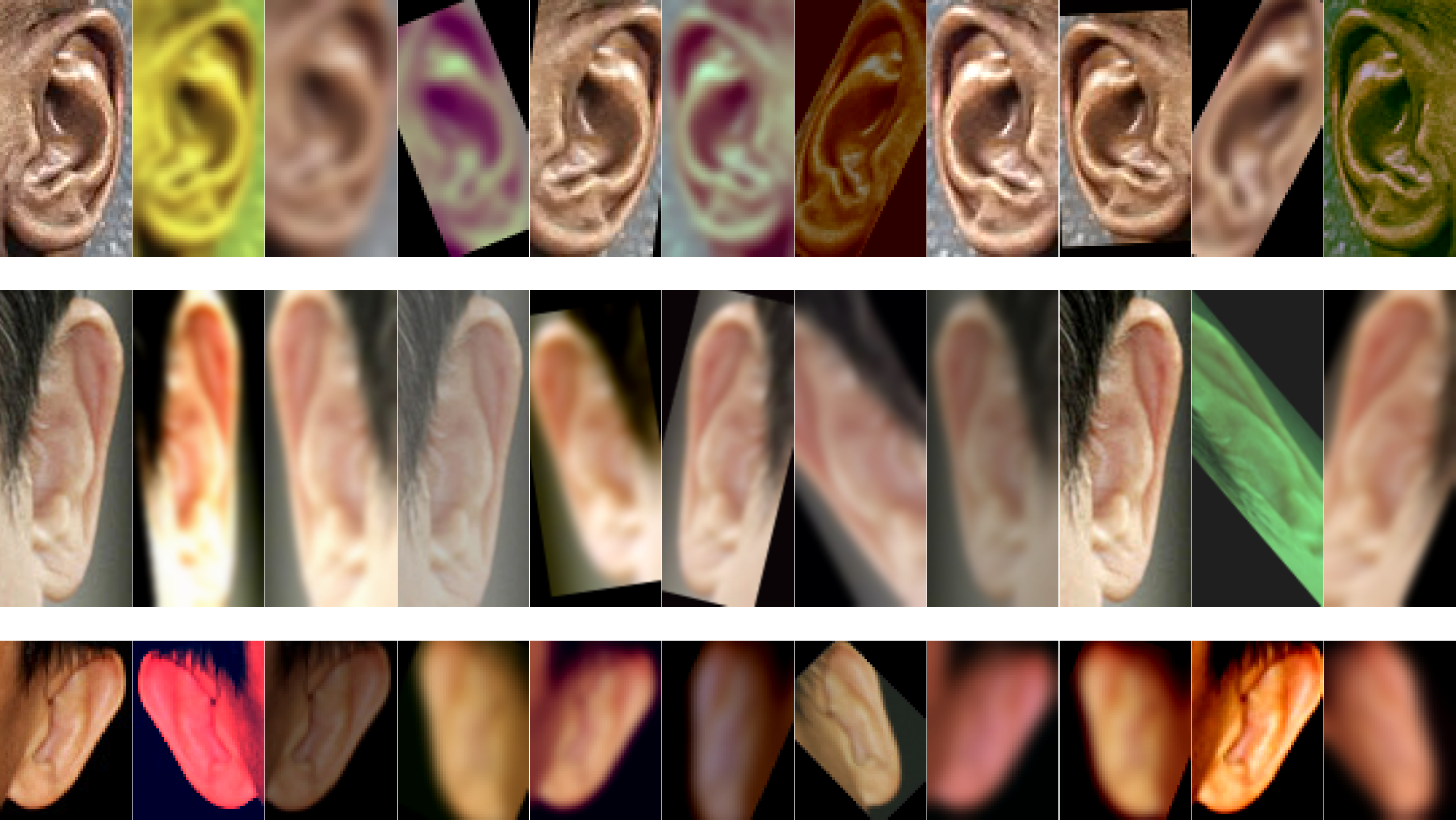}	
	\caption{Augmentation examples: the left most image in each row represents the original image from our dataset, while the following 10 images are the augmented variations. 
%NIMA SKLICA V TEKSTU + SLIKA MORA BITI ZA SLIKO 2, ALI PA ENOSTAVNO TUKAJ POKAZES SE NA TE TRI SLIKE IZ BAZE ... IN SLIKE 2 NE RABIŠ ... TODO!!!
Note the extend of variability added to our training data by the augmented images.}
		\label{figure:augmentations}
\end{figure*}
As part of our \textit{selective model learning} strategy, we therefore initialize each of the considered architectures with parameters learned on the ImageNet dataset and fine-tune certain layers only. The exception here is the last fully-connected layer that needs to be learned from scratch and the softmax classifier at the top of the network that provides the output for our closed-set recognition problem. Below, we outline details of the selective training for each considered model architecture:
\begin{itemize}
 \item AlexNet and VGG-16: For both architecture, we fix all models layers learned on the ImageNet data and initialize the first two fully connected layers with weights learned on ImageNet. Using the available training data for our problem, we fine tune these two layers and learn the third fully connected layer (which represents a proxy for the application-dependent softmax classifier) from scratch. %The learning rate for the first two model layers is set to a smaller vcompared to FC8 that was learned from scratch.
 \item{SqueezeNet}: Since this model has no fully-connected layers, we fine tune all layers except the last convolutional layer that has a different number of classes and therefore requires full learning. All model parameter are again initialized using the ImageNet data.
 \end{itemize}

\subsection{Data augmentation}

Both of our training strategies -- full model learning as well as selective learning -- require a sufficient amount of  data for the training procedure~\cite{donahue2014decaf}. 
%This strategy tests different levels of augmentations. CNN require large amounts of data to learn successfully~\cite{donahue2014decaf}. 
In order to satisfy this requirement, we perform augmentations of the original dataset with translations, mild rotations, and color variations using the Imgaug tool ({\small\url{http://github.com/aleju/imgaug}}), where the data-transformations are performed (or not) with a 50\% chance. Below is  a list of the augmentation procedures we used to increase the amount of available training data.
\begin{itemize}
\item horizontal flipping,
\item trimming 0--10\% of images on each side,
\item Gaussian blurring with $\sigma$ 0--3.0,
\item addition of Gaussian noise with scale 0--0.2,
\item brightness reduction/increase of pixel intensities by a value of 10 (over all color channels or over a single channel),
\item contrast increase/decrease of up to 50\% (over all color channels or over a single channel),
\item rotation of up to 45$^{\circ}$ in both directions,
\item scale increase/decrease of up to 20\%.
\end{itemize}
Some sample augmentations are shown in Fig.~\ref{figure:augmentations}. Here, the left most image in each row shows an example of an original image from our dataset, while the rest represents synthetic augmentations.

% Motivacija, povej tukaj kaj smo pri katerem modelu učili, če lahko oceniš število parametrov še toliko bolje. Izpusti stvari, ki niso delale.
%Furthermore, we report the results on different perturbation levels and thus enhancing performance even further.

% I

%%%%%%%%%%%%%%%%%%%%%%%%%%%%%%%%%%%%%%%%%%%%%%%%%%%%%%%%%%%%%%%%%%%%%%%%%%%%%%%%%
\section{Experimental setup and results}
\label{section:experiments}

In our experiments we report the results for all strategies to CNN-model training outlined in the previous section. We present results for full-model and selective-model learning for all three architectures, but explore the impact of aggressive data augmentation only for the model with the least amount of parameters, i.e., SqueezeNet. % performing architecture. %It also needs to be noted that during all experiments, all left ear were flipped%In the latter different amounts of augmentations are used: none, $10\times$ and $100\times$. However, because the highest amount of augmentations proved to be the most successful, we report only the $100\times$ augmentations in all other experiments -- namely, full model learning and selective model learning experiments were performed on the $100\times$ augmented datasets.

\subsection{Dataset} %%%%%%%%%%%%%%%%%%%%%%%%%%%%%%%%%%%%%%%%%%%%%%%%%%%%%%%%%%%%%%%%%%%%%%%%%%%%%%%%%

For our experiments we compose a dataset of unconstrained ear images by merging the recently introduced AWED and CVLED datasets~\cite{awe}. To have more data available to work with, we collect an additional 
500 images of 50 subject from the Internet.
%... ZDEJ LAHKO POVEŠ KAJ O KARAKTERISTIKAH, ... 
%The dataset was annotated according to the following categories: gender, ethnicity, accessories, occlusion, head pitch, head roll, head yaw, and head side.
The combined dataset contains 2304 images (1000 from AWED, 804 from CVLED, and 500 newly collected) of 166 persons.

\begin{figure}[tb]
\captionsetup{type=figure}
	\center
		\subfloat{
			\includegraphics[height=0.4\columnwidth]{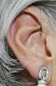}
		}
		\subfloat{
			\includegraphics[height=0.4\columnwidth]{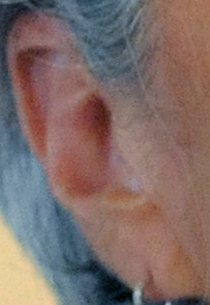}
		}
%		\subfloat{
% 			\includegraphics[height=0.2\columnwidth]{awed_samples/p1-3}
% 		}
% 		\subfloat{
% 			\includegraphics[height=0.2\columnwidth]{awed_samples/p1-4}
% 	}
% 	\subfloat{
% 			\includegraphics[height=0.2\columnwidth]{awed_samples/p1-5}
% 		}
		\subfloat{
			\includegraphics[height=0.4\columnwidth]{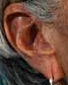}
		}\\
		\subfloat{
			\includegraphics[height=0.4\columnwidth]{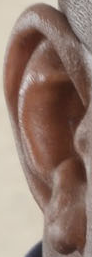}
		}
		\subfloat{
			\includegraphics[height=0.4\columnwidth]{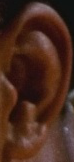}
		}
% 		\subfloat{
% 			\includegraphics[height=0.4\columnwidth]{awed_samples/p2-3}
% 		}
% 		\subfloat{
% 			\includegraphics[height=0.2\columnwidth]{awed_samples/p2-4}
% 		}
% 		\subfloat{
% 			\includegraphics[height=0.2\columnwidth]{awed_samples/p2-5}
% 		}
 		\subfloat{
 			\includegraphics[height=0.4\columnwidth]{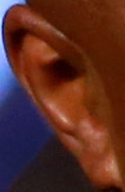}
 		}
		\subfloat{
			\includegraphics[height=0.4\columnwidth]{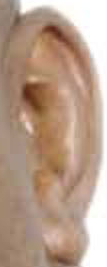}
		}
		%\subfloat{
		%	\includegraphics[height=0.3\columnwidth]{awed_samples/p2-8}
		%}
		\caption{Sample images from the dataset. Images in each row correspond to one subject from the dataset. Note the extend of variability present in the images.}
		\label{figure:awed_sample}
\end{figure}

%\textbf{Acquisition and structure:}
%The images were collected with two different strategies:
%\begin{itemize}
%\item
All images are collected in a consistent manner from the internet using web crawlers and a subsequent manual inspection. Since the main purpose of the collected images is not ear-recognition research, the images contain realistic variations and present a challenging task to automatic ear recognition technology. Variations that can be found in the dataset are across gender, head rotations, race, presence of occlusions and alike.   %that sent appropriate search queries to Google's and Bing's image search. The collected images were then manually screened and only a selected number of images per subject were selected for inclusion in the dataset. 
%Some sample images are shown in Fig.~\ref{figure:awed_sample} GLEJ CAPTION 2 TODO!!!.
A few sample images from the dataset are shown in Figs.~\ref{figure:augmentations} in \ref{figure:awed_sample}.

%\item Additional images were taken from the LFW dataset~\cite{lfwtodo} as impostors. Ears were automatically segmented out of the face images using Haar-cascade detector~\cite{haartodo} with some presumptions regarding minimal sized of detected areas.% todo: opisi tocno kaj so bile te predpostavke in zakaj smo jih dali! % opisi se tudi zakaj so impostorji pomembni! tole se je treba se razpisat
%A trained researcher then manually checked and removed unwanted images were ears were not detected correctly or contained ears from other persons. Furthermore, all images from persons already present in our dataset acquired from the internet, were removed as well. This procedure resulted in 7700 images of 3360 persons, whereas LFW originally contains 13233 images of 5749 people.
%\end{itemize}

 %The test part of the dataset contains 9,500 ear images of 3,540 subjects, 7,700 (3360 subjects) of those represent images of impostors. The images are tightly cropped and do not contain profile faces as shown in Fig.~\ref{figure:awed_sample}.
% TODO tole popravi zgoraj
% mogoce bi tu lahko dodali diagram kako je baza dejansko sestavljena? - kateri del je iz AWE, kateri iz CVLEDB, katero je geniune del, katero impostors.

%1500 images of 150 subjects in the dataset was annotated according to the following categories: gender, ethnicity, accessories, occlusion, head pitch, head roll, head yaw, and head side. The pitch, roll and yaw angles were estimated from the head pose. 
%The available labels for each category are shown in Table~\ref{Tab:_labels} and the distribution of the labels for the entire dataset in Fig.~\ref{img:db_1000}. 
%Additionally, the location of the tragus was also marked in each image.

\subsection{Performance metrics \& protocols} %%%%%%%%%%%%%%%%%%%%%%%%%%%%%%%%%%%%%%%%%%%%%%%%%%%%%%%%%%%%%%%%%%%%%%%%%%%%%%%%%

We perform identification experiments with a closed set experimental protocol. This means that for each image in the dataset the CNN-based recognition model needs to predict to which of the 166 classes the input image corresponds to. Based on that we report the following performance curves and metrics for the identification experiments:

\begin{itemize}
  \item Cumulative Match-score Curves (CMC),
  \item Rank-1 and Rank-5 recognition rates,
  \item Area Under the CMC Curve (AUCMC).
\end{itemize}
% damu tu se kaksno referenco

For the experiments, the dataset is split into train and test sets in a ratio of 60\% vs. 40\%, respectively. Splitting is done for each subject separately: 60\% of images of a given subject are randomly selected for the train set, the rest for the test set. This means that in the train set there are 1,383 images and 921 are in the test set. All results are reported in identification experiments with the 921 test images. %However, the images were augmented as specified in Section~\ref{section:methodology}. 

%For the data augmentation we only used images from the train set. In the experiments we asWe considered different levels of data augmentations This means that the effective number of images was 13,830 with 10$\times$ augmentations for each image and 138,300 with 100$\times$ augmentations per image, instead of 1,383 images that were originally present in the train set.

\subsection{Results} %%%%%%%%%%%%%%%%%%%%%%%%%%%%%%%%%%%%%%%%%%%%%%%%%%%%%%%%%%%%%%%%%%%%%%%%%%%%%%%%%
\begin{table}
%\normalsize
\centering
\caption{Performance metrics showing the effect of 
%perturbations 
data augmentation on the recognition performance. Results here are shown for the SqueezeNet architecture only. Note how the increase in available training data improves performance.}
\label{resultsAug}
\begin{tabular}{@{}lllll@{}}
\toprule
\#Iter. & \#Augm. & Rank-1 [\%] & Rank-5 [\%] & AUCMC [\%]  \\ \midrule
10k & 0     & 36.37	&	57.33	&	89.91	\\
    & 10    & 46.04	&	69.49	&	94.10	\\
    & 100   & 56.68	&	77.09	&	96.09	\\ \midrule
20k & 0    	& 40.72	&	61.45	&	90.93	\\
    & 10    & 56.57	&	74.92	&	93.88	\\
    & 100   & 61.67	&	79.26	&	95.62	\\\midrule
30k & 0     & 41.26	&	61.45	&	90.96	\\
    & 10    & 56.89	&	75.24	&	93.85	\\
    & 100   & 61.89	&	80.46	&	95.48	\\\midrule
40k & 0     & 41.26	&	61.45	&	90.96	\\
    & 10    & 56.89	&	75.46	&	93.86	\\
    & 100   & 62.00	&	80.35	&	95.51	\\\midrule
50k & 0     & 41.26	&	61.45	&	90.96	\\
    & 10    & 57.00	&	75.46	&	93.86	\\
    & 100   & 62.00	&	80.35	&	95.51	\\
\end{tabular}
\end{table}

\begin{figure}[!ht!]
    \begin{center}
        \includegraphics[width=0.99\columnwidth,trim=0.5cm 6cm 1cm 6.5cm,clip]{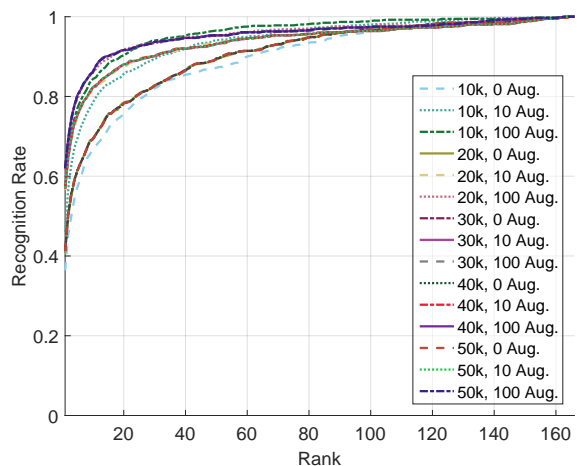}
        \caption{CMC curves showing the effect of data augmentation on the recognition performance.  Here we show only results for the SqueezeNet architecture. The figure legend provides information about the number of training iterations  and augmentation factor used, i.e., (\#Iter, factor Aug.).}
        \label{test5}
        \vspace{-4mm}
    \end{center}
\end{figure}

\begin{table*}
\caption{Performance metrics for selective learning (bottom) and full model learning (top). The tables show results for: different model architectures: (a) AlexNet, (b) VGG-16, and (c) SqueezeNet. The best results for each model architecture are presented in italic and the best overall results are marked bold. %The best performing model is SqueezNet trained with the selective model learning strategy and aggressive data augmentation.
}
\label{TabulatedResults}
\begin{minipage}{0.32\textwidth}
\centering
\scriptsize
\caption*{\footnotesize (a) AlexNet results}
%\label{resultsSqueezeNet}
\begin{tabular}{@{}llll@{}}
\toprule
\# Iter. & Rank-1 [\%] & Rank-5 [\%] & AUCMC [\%]  \\ \midrule
\multicolumn{4}{c}{Full model learning} \\ \midrule
10k	&	34.85	&	53.31	&	88.74	\\
20k	&	36.16	&	52.66	&	88.89	\\
30k	&	37.57	&	55.48	&	89.19	\\
40k	&	37.24	&	55.48	&	89.16	\\
50k	&	37.46	&	55.37	&	89.08	\\ \midrule
\multicolumn{4}{c}{Selective model learning} \\ \midrule
10k	&	46.15	&	67.32	&	94.04	\\
20k	&	49.29	&	69.60	&	94.48	\\
30k	&	49.19	&	\textit{69.92}	&	94.55	\\
40k	&	\textit{49.51}	&	69.71	&	94.54	\\
50k	&	\textit{49.51}	&	69.82	&	\textit{94.57}	\\
\end{tabular}
\end{minipage}
\begin{minipage}{0.32\textwidth}
\centering
\caption*{\footnotesize (b) VGG-16 results}
%\label{resultsSqueezeNet}
\scriptsize
\begin{tabular}{@{}llll@{}}
\toprule
\# Iter. & Rank-1 [\%] & Rank-5 [\%] & AUCMC [\%]  \\ \midrule
\multicolumn{4}{c}{Full model learning} \\ \midrule
10k	&	32.14	&	52.88	&	89.23	\\
20k	&	43.87	&	62.00	&	92.63	\\
30k	&	46.25	&	64.39	&	92.87	\\
40k	&	46.36	&	64.60	&	92.16	\\
50k	&	49.08	&	66.67	&	92.99	\\ \midrule
\multicolumn{4}{c}{Selective model learning} \\ \midrule
10k	&	48.10	&	67.97	&	94.14	\\
20k	&	49.19	&	70.03	&	94.45	\\
30k	&	50.27	&	70.90	&	94.66	\\
40k	&	51.14	&	71.77	&	94.78	\\
50k	&	\textit{51.25}	&	\textit{71.99}	&	\textit{94.81}	\\
\end{tabular}
\end{minipage}
\begin{minipage}{0.32\textwidth}
\scriptsize
\centering
\caption*{\footnotesize (a) SqueezeNet results}
%\label{resultsSqueezeNet}
\begin{tabular}{@{}llll@{}}
\toprule
\# Iter. & Rank-1 [\%] & Rank-5 [\%] & AUCMC [\%]  \\ \midrule
\multicolumn{4}{c}{Full model learning} \\ \midrule
10k	&	22.15	&	42.24	&	85.43	\\
20k	&	31.38	&	51.25	&	88.97	\\
30k	&	35.07	&	55.48	&	88.87	\\
40k	&	36.81	&	55.92	&	88.05	\\
50k	&	36.92	&	56.03	&	87.58	\\ \midrule
\multicolumn{4}{c}{Selective model learning} \\ \midrule
10k	&	56.68	&	77.09	&	\textbf{96.09}	\\
20k	&	61.67	&	79.26	&	95.62	\\
30k	&	61.89	&	\textbf{80.46}	&	95.48	\\
40k	&	\textbf{62.00}	&	80.35	&	95.51	\\
50k	&	\textbf{62.00}	&	80.35	&	95.51	\\
\end{tabular}
\end{minipage}
\end{table*}
\begin{figure*}
\begin{minipage}{0.33\textwidth}
  \centering
  \includegraphics[width=1\textwidth,trim=0.5cm 6cm 0cm 6cm,clip]{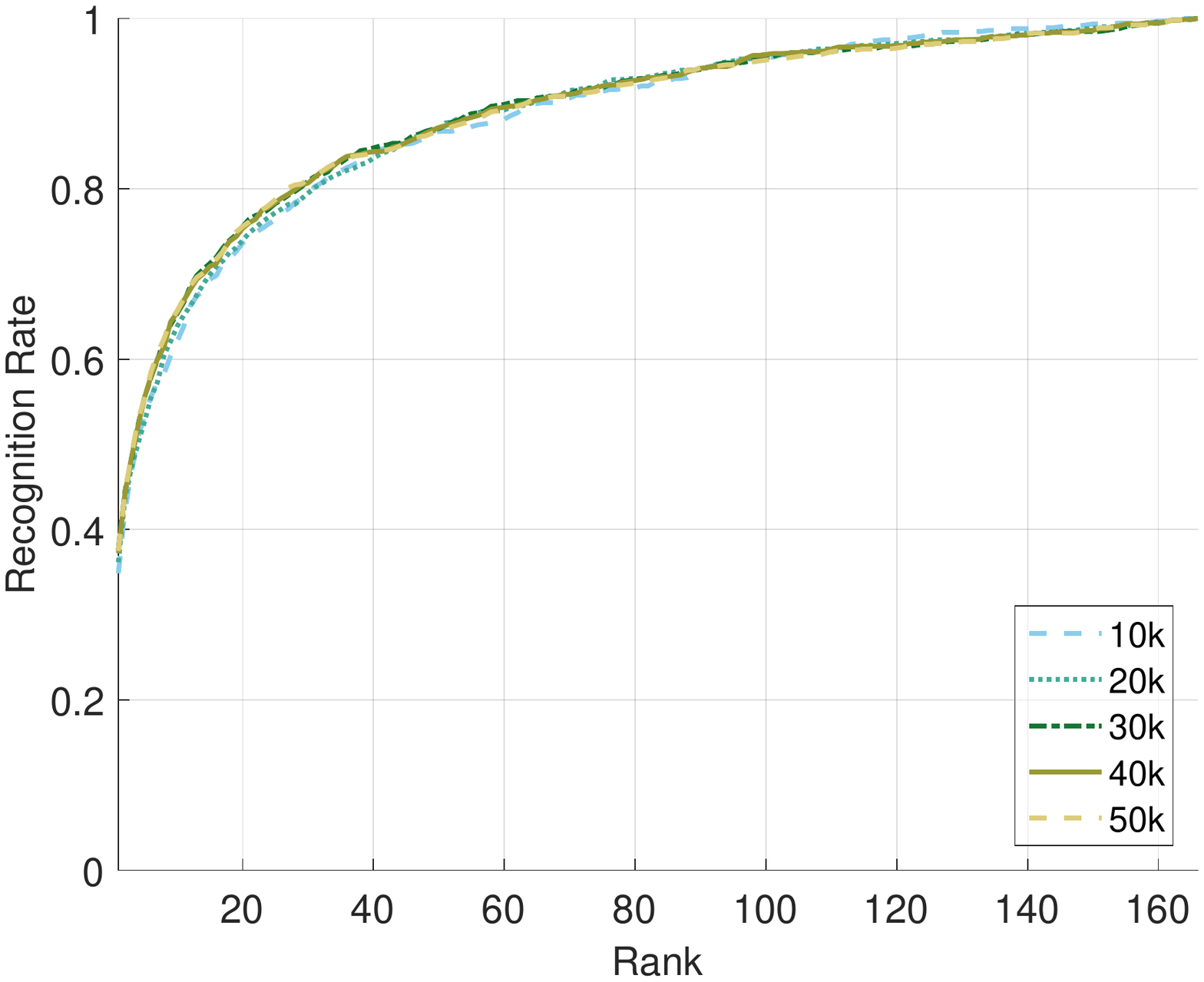}
  \caption*{(a) AlexNet -- full model learning}
  \label{test1a}
\end{minipage}
\begin{minipage}{0.33\textwidth}
  \centering
  \includegraphics[width=1\textwidth,trim=0.5cm 6cm 0cm 6cm,clip]{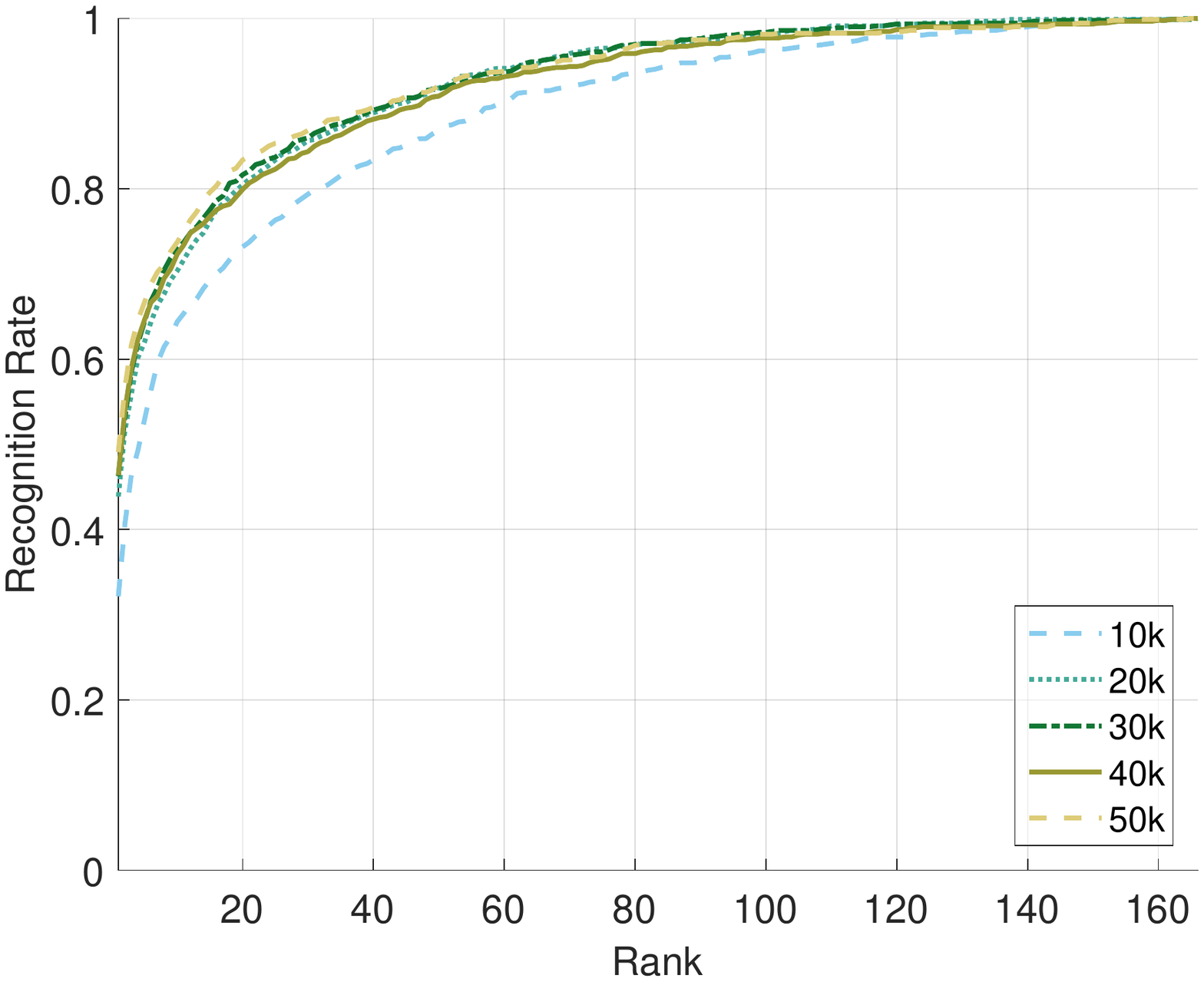}
  \caption*{(b) VGG -- full model learning}
  \label{test2a}
\end{minipage}
\begin{minipage}{0.33\textwidth}
  \centering
  \includegraphics[width=1\textwidth,trim=0.5cm 6cm 0cm 6cm,clip]{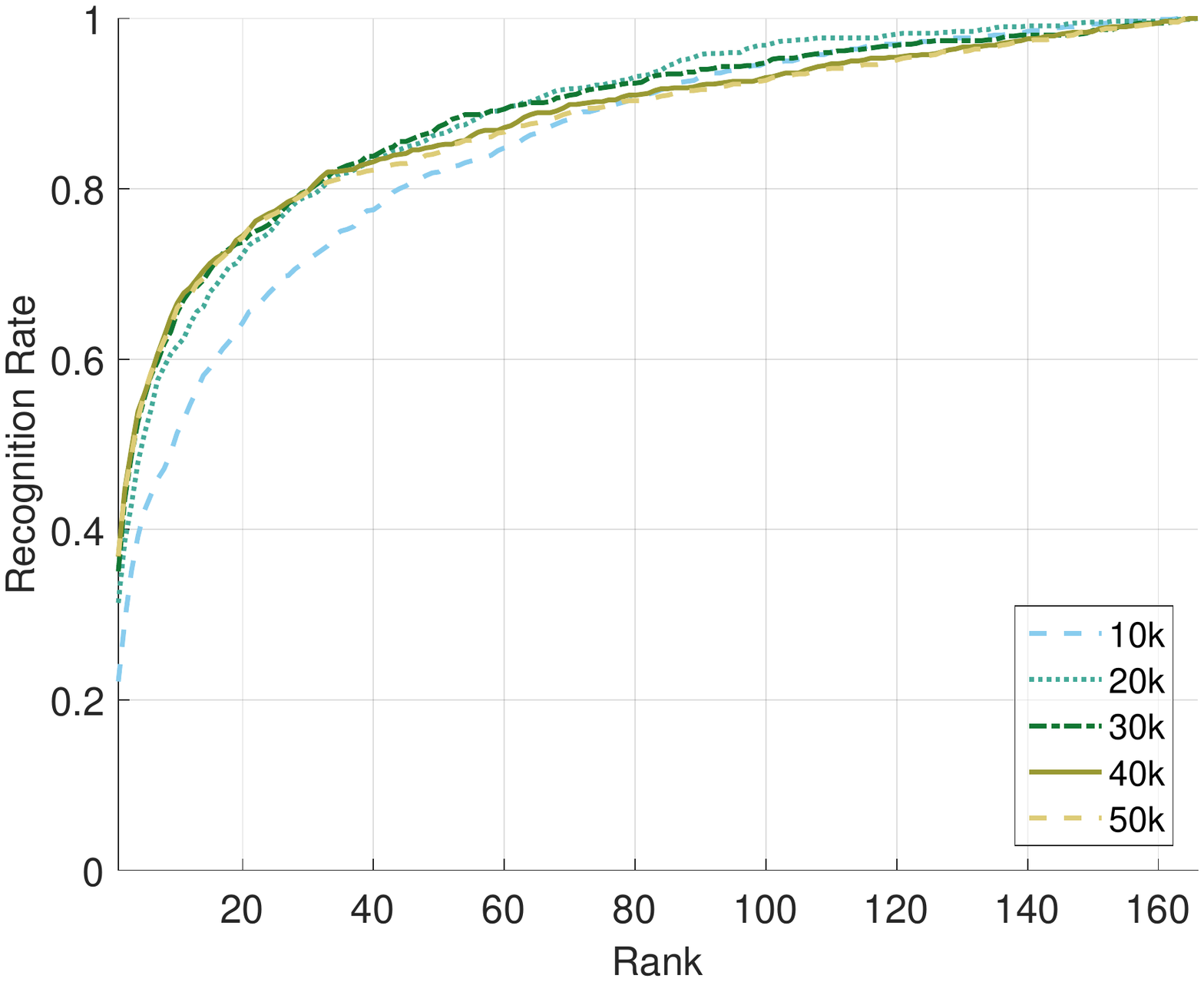}
  \caption*{(c) SqueezeNet -- full model learning}
  \label{test3a}
\end{minipage}

\begin{minipage}{0.33\textwidth}
  \centering
  \includegraphics[width=1\textwidth,trim=0.5cm 6cm 0cm 6cm,clip]{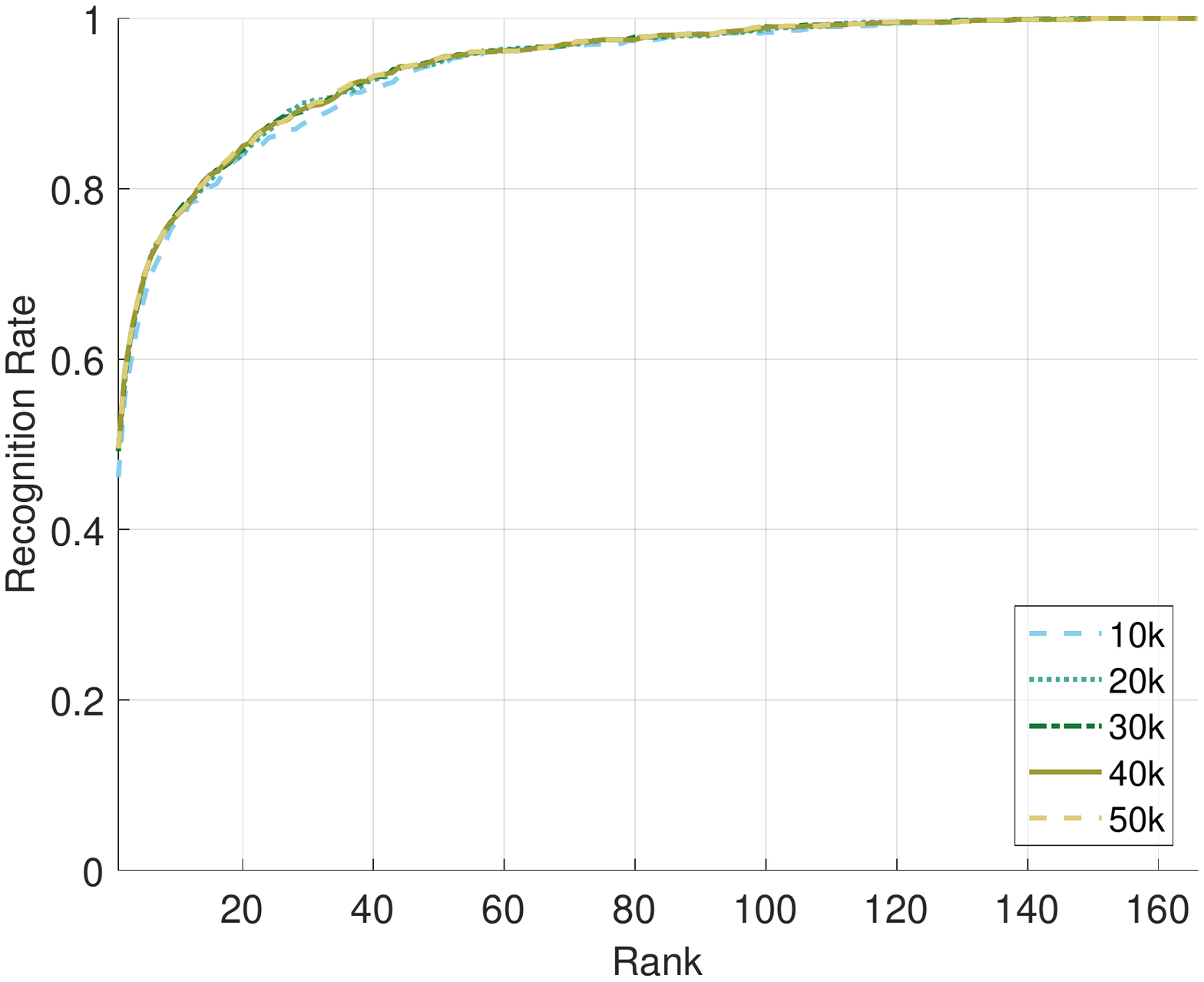}
  \caption*{(d) AlexNet -- selective model learning}
  \label{test1b}
\end{minipage}
\begin{minipage}{0.33\textwidth}
  \centering
  \includegraphics[width=1\textwidth,trim=0.5cm 6cm 0cm 6cm,clip]{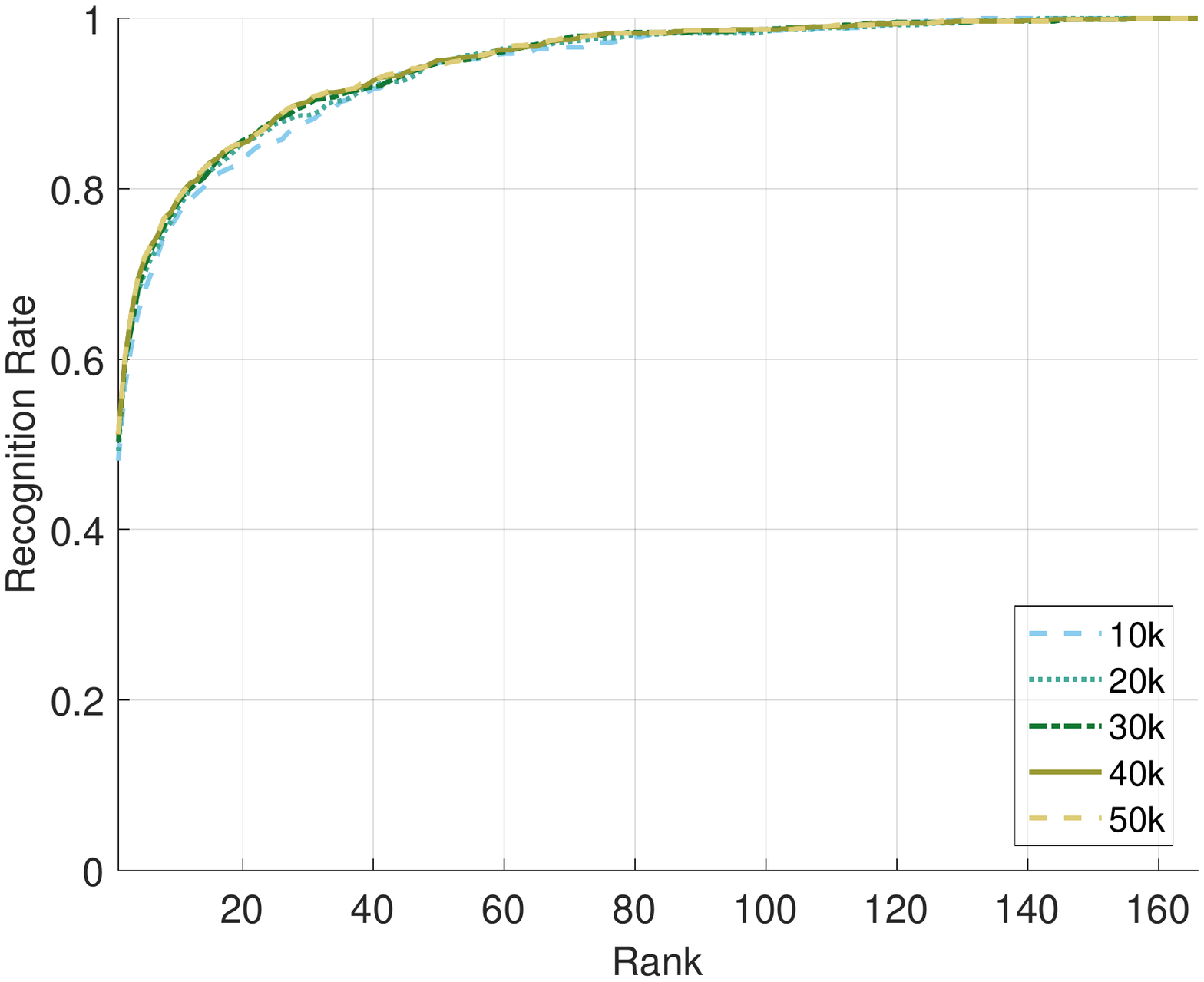}
  \caption*{(e) VGG -- selective model learning}
  \label{test2b}
\end{minipage}
\begin{minipage}{0.33\textwidth}
  \centering
  \includegraphics[width=1\textwidth,trim=0.5cm 6cm 0cm 6cm,clip]{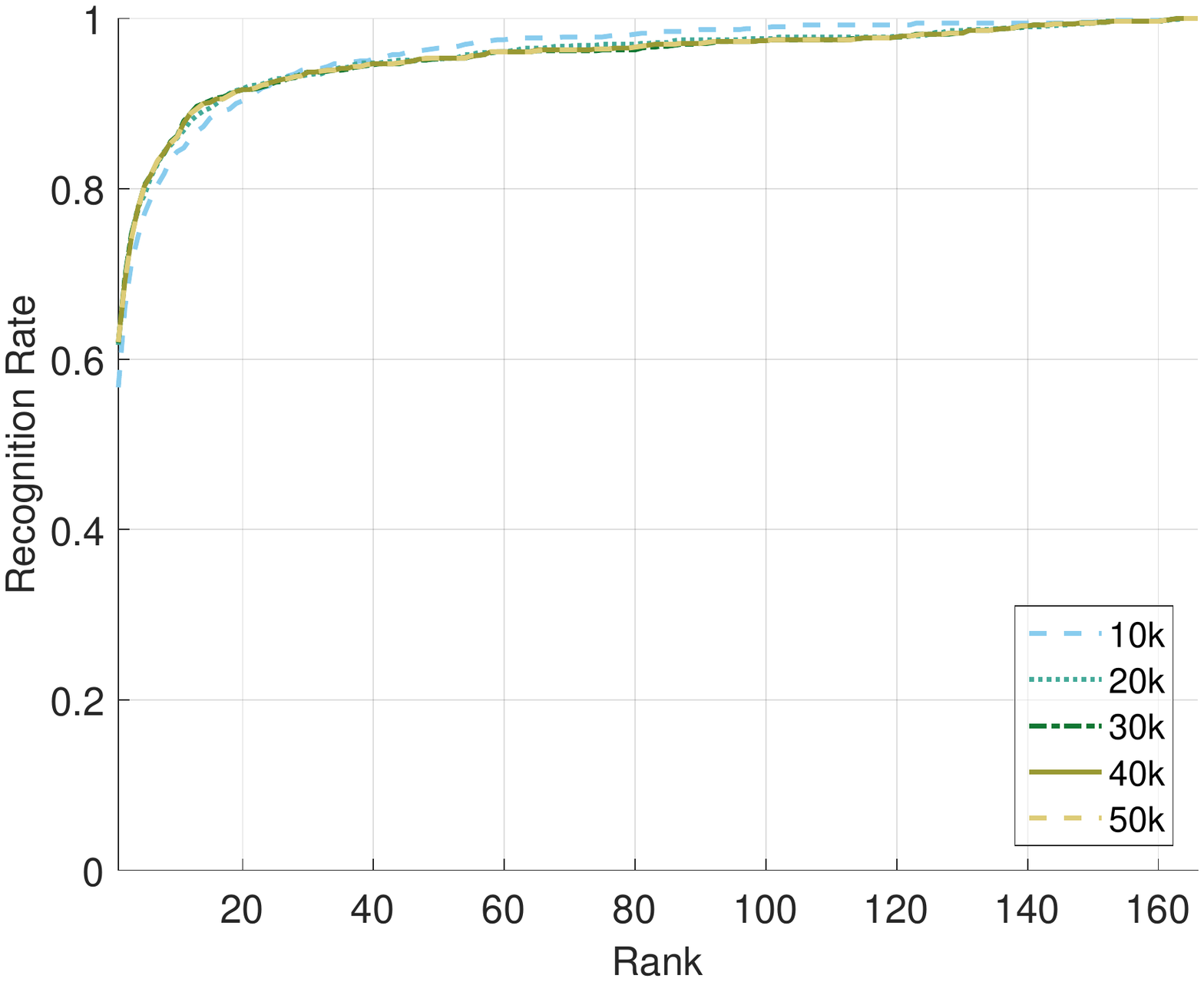}
  \caption*{(f) SqueezeNet -- selective model learning}
  \label{test3b}
\end{minipage}
\caption{CMC curves generated based on different learning strategies. The first column shows curves for the AlexNet architecture, the second column shows results for the VGG architecture and the third column shows results for the SqueezeNet architecture. The upper row depicts results for full model learning and the lower for selective model learning. The best results are obtained with the SqueezeNet architecture with selective model learning. The results are best viewed in color.}
\label{tests1to3}
\vspace{-4mm}
\end{figure*}

In the first series of experiments we  evaluate the impact of aggressive data augmentation on CNN-based model training. For these experiments we consider only the SqueezeNet architecture, which is the fastest to train due to the relatively small number of model parameters. We perform experiments without data augmentation with the initial set of 1383 training images (i.e., an augmentation factor of 0), with 10-times the original training data (i.e., an augmentation factor of 10) and with 100-times the original training data (i.e., an augmentation factor of 100). We observe the recognition performance of SqueezeNet trained with selective learning after 10,000, 20,000, 30,000, 40,000 and 50,000 training iterations. The results on the test set are presented in Table~\ref{resultsAug} and Fig.~\ref{test5}. After each group of iterations the augmentations prove to be very effective, increasing Rank-1 recognition rate by more than 20 percentage points when comparing no augmentation and the augmentation with a factor of 100. As expected, the highest augmentation factor results in the best overall performance. We, therefore, report all following results for experiments with an augmentation factor of 100. %We also see that the performance does not change much after 20000 iterations.  

In the next series of experiments, we explore our training strategies on all three model architectures. We fix the data augmentation factor to 100 for all tests and vary only the number of training iterations from 20,000 to 50,000. The results for AlexNet in Table~\ref{TabulatedResults}(a) %{resultsAlexNet} 
show that performance steadily increases as the number of iterations increases and starts stagnating after 30,000 iterations. However, while the trend is the same for both of our training strategies (full model learning and selective model learning) selective model learning significantly outperforms the full model learning approach with a 49.51\% Rank-1 recognition rate vs. 37.46\% for full model learning. The same conclusion holds for the Rank-5 and AUCMC performance metrics with 69.82\% vs. 55.37\% and 94.57\% vs. 89.08\%, respectively. The two plots in the first column of Fig.~\ref{tests1to3} show CMC curves and again support the conclusion, where the CMC curves for selective model learning are higher for all iterations.
% \begin{table}
% %\normalsize
% \centering
% \caption{AlexNet performance metrics for learning on fully-connected layers (bottom) and full model learning (top).}
% \label{resultsAlexNet}
% \begin{tabular}{@{}llll@{}}
% \toprule
% \# Iter. & Rank-1 [\%] & Rank-5 [\%] & AUCMC [\%]  \\ \midrule
% \multicolumn{4}{c}{Full model learning} \\ \midrule
% 10k	&	34.85	&	53.31	&	88.74	\\
% 20k	&	36.16	&	52.66	&	88.89	\\
% 30k	&	37.57	&	55.48	&	89.19	\\
% 40k	&	37.24	&	55.48	&	89.16	\\
% 50k	&	37.46	&	55.37	&	89.08	\\ \midrule
% \multicolumn{4}{c}{Selective model learning} \\ \midrule
% 10k	&	46.15	&	67.32	&	94.04	\\
% 20k	&	49.29	&	69.60	&	94.48	\\
% 30k	&	49.19	&	69.92	&	94.55	\\
% 40k	&	49.51	&	69.71	&	94.54	\\
% 50k	&	49.51	&	69.82	&	94.57	\\
% \end{tabular}
% \end{table}

For VGG-16 the selective model learning again outperforms full model learning, as shown in Table~\ref{TabulatedResults}(b)  
%{resultsVGG} 
and the two plots in the middle column of Fig.~\ref{tests1to3}, but by a small margin of 2.17 percentage points for the Rank-1 recognition rate. The selective model learning on VGG-16 outperforms AlexNet with a Rank-1 recognition rate of 51.25\% compared to the 49.51\% of AlexNet. The best Rank-5 recognition rate and AUCMC values for the VGG-16 model are 71.99\% and 94.81\%, respectively, and are higher than with AlexNet.

% \begin{table}
% %\normalsize
% \centering
% \caption{VGG-16 performance metrics for learning on fully-connected layers (bottom) and full model learning (top).}
% \label{resultsVGG}
% \begin{tabular}{@{}llll@{}}
% \toprule
% \# Iter. & Rank-1 [\%] & Rank-5 [\%] & AUCMC [\%]  \\ \midrule
% \multicolumn{4}{c}{Full model learning} \\ \midrule
% 10k	&	32.14	&	52.88	&	89.23	\\
% 20k	&	43.87	&	62.00	&	92.63	\\
% 30k	&	46.25	&	64.39	&	92.87	\\
% 40k	&	46.36	&	64.60	&	92.16	\\
% 50k	&	49.08	&	66.67	&	92.99	\\ \midrule
% \multicolumn{4}{c}{Selective model learning} \\ \midrule
% 10k	&	48.10	&	67.97	&	94.14	\\
% 20k	&	49.19	&	70.03	&	94.45	\\
% 30k	&	50.27	&	70.90	&	94.66	\\
% 40k	&	51.14	&	71.77	&	94.78	\\
% 50k	&	51.25	&	71.99	&	94.81	\\
% \end{tabular}
% \end{table}

SqueezeNet, same as AlexNet and VGG-16, performs better with selective model learning than full model learning, which again suggests that good parameter initialization is crucial for model training and helps reduce the need for large amounts of training data. The results for the SqueezeNet architecture are shown in Table~\ref{TabulatedResults}(c)  
%{resultsSqueezeNet} 
and the CMC plots in the last column of Fig.~\ref{tests1to3}. If we look at the  results for full model learning only, SqueezeNet does not outperform VGG-16 with a Rank-1 recognition rate of 36.92\%  compared to 49.08\% for VGG-16. However, it outperforms both, AlexNet and VGG-16, under the selective model learning strategy. The model results in a Rank-1 recognition rate of 62.00\%  vs. 49.51\% and 51.25\% for AlexNet and VGG-16, respectively.

Overall, the results suggest that selective model learning should be preferred to full model learning when only limited training data is available. While the results obtained with full-model learning were better than we initially expected, the training procedure proved much more difficult than with selective model learning. To find a good parameter configuration for the considered architectures with full model learning, we needed to use hyper-parameter optimization, test different parameter initialization schemes and repeat the training procedure several times. With selective model learning, on the other hand, the training procedure was straight forward and quickly converged to a good model configuration. 
%\begin{table}
%\normalsize
%\centering
%\caption{SqueezeNet performance metrics for selective learning (bottom) and full model learning (top).}
%\label{resultsSqueezeNet}
%\begin{tabular}{@{}llll@{}}
%\toprule
%\# Iter. & Rank-1 [\%] & Rank-5 [\%] & AUCMC [\%]  \\ \midrule
%\multicolumn{4}{c}{Full model learning} \\ \midrule
%10k	&	22.15	&	42.24	&	85.43	\\
%20k	&	31.38	&	51.25	&	88.97	\\
%30k	&	35.07	&	55.48	&	88.87	\\
%40k	&	36.81	&	55.92	&	88.05	\\
%50k	&	36.92	&	56.03	&	87.58	\\ \midrule
%\multicolumn{4}{c}{Selective model learning} \\ \midrule
%10k	&	56.68	&	77.09	&	96.09	\\
%20k	&	61.67	&	79.26	&	95.62	\\
%30k	&	61.89	&	80.46	&	95.48	\\
%40k	&	62.00	&	80.35	&	95.51	\\
%50k	&	62.00	&	80.35	&	95.51	\\
%\end{tabular}
%\end{table}

%We also performed aggressive data augmentation. With our train set containing only 1,383 images, there was a great promise for data augmentation to improve performance. 

\begin{figure}
    \begin{center}
        \includegraphics[width=0.99\columnwidth,trim=1cm 6cm 1.6cm 6cm,clip]{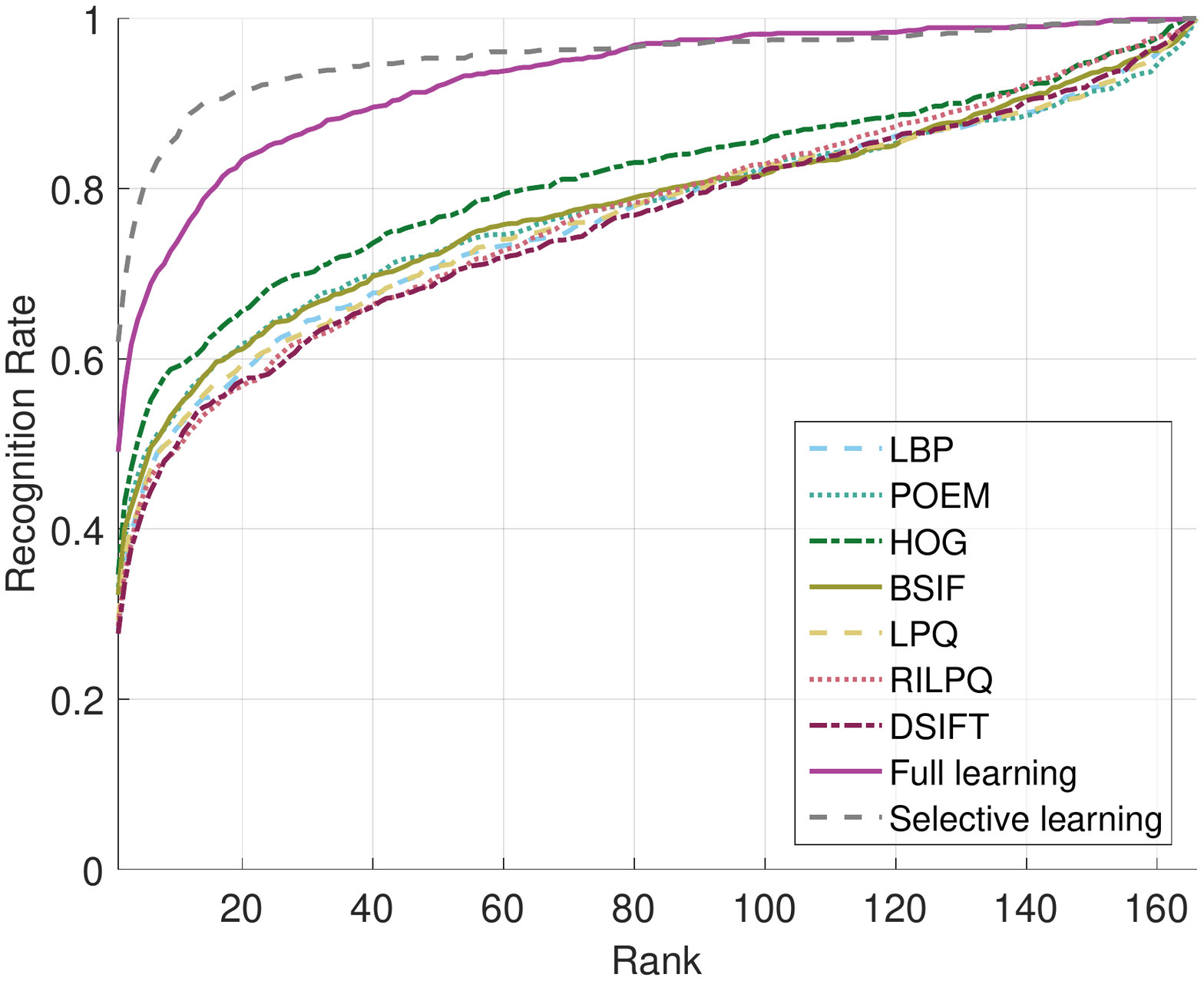}
        \caption{CMC curves comparing state-of-the-art feature extraction methods to our best performing CNN-based approaches for full model learning (VGG-16) and for selective model learning (SqueezeNet).}
        \label{test4}
        \vspace{-4mm}
    \end{center}
\end{figure}

In the last series of identification experiments, we compare some existing state-of-the-art techniques to the best architectures obtained with the full-model-learning and selective-model-learning strategies. Here, we select the techniques to be included in our experiments based on recent publications evaluating the performance of various ear recognition techniques, i.e.~\cite{awe,anika}. The state-of-the-art techniques included in our experiments are local  techniques implemented in the AWE toolbox~\cite{awe}, which exploit the following descriptors: LBP~\cite{guo2008ear}, POEM~\cite{vu2010face}, HOG~\cite{damer2012ear}, BSIF~\cite{kannala2012bsif}, LPQ~\cite{ojansivu2008blur}, RILPQ~\cite{ojansivu2008rotation}, and DSIFT~\cite{dewi2006ear}. The results of this series of experiments are presented in Table~\ref{resultsAll} and Fig.~\ref{test4}. As can be seen, even the fully learned VGG-16 model significantly outperforms the best performing local approach, i.e., based on the Histogram-of-Oriented-Gradients (HOG) descriptor, with 49.08\% vs. 34.64\%, respectively. Furthermore, with selective learning SqueezeNet outperforms both HOG and POEM (which are the two best performing local techniques in our experiments) by a significant margin: 62.00\% Rank-1 recognition rate compared to 34.64\% for HOG and 33.12\% for POEM. The same holds for Rank-5 recognition rates 92.99\% and AUCMC 95.51\% compared to 52.88\% and 81.07\% for HOG and 48.53\% and 77.34\% for POEM. %The results are shown in Table~\ref{resultsAll} and Fig.~\ref{test4}.

\begin{table}
%\normalsize
\centering
\caption{Performance metrics for the top performing CNN-based approaches for the full-model-learning and for the selective-model-learning strategies vs. some of the state-of-the-art feature extraction approaches. Results for the top performing local techniques are presented in italic, the overall best performance is marked bold.}
\label{resultsAll}
\begin{tabular}{@{}lrrr@{}}
\toprule
\# Iter. & Rank-1 [\%] & Rank-5 [\%] & AUCMC [\%]  \\ \midrule
LBP~\cite{guo2008ear}	&	28.12	&	45.17	&	76.40	\\
POEM~\cite{vu2010face}	&	33.12	&	48.53	&	77.34	\\
HOG~\cite{damer2012ear}	&	\textit{34.64}	&	\textit{52.88}	&	\textit{81.07}	\\
BSIF~\cite{kannala2012bsif}	&	32.25	&	47.23	&	77.69	\\
LPQ~\cite{ojansivu2008blur}	&	29.21	&	44.95	&	76.49	\\
RILPQ~\cite{ojansivu2008rotation}	&	28.56	&	44.41	&	76.76	\\
DSIFT~\cite{dewi2006ear}	&	27.69	&	42.67	&	75.74	\\
Full learning (VGG-16)	&	49.08	&	66.67	&	92.99	\\
Sel. learning (SqueezeNet)	&	\textbf{62.00}	&	\textbf{80.35}	&	\textbf{95.51}	\\
\end{tabular}
\end{table}

\section{Conclusion}
\label{section:conclusion}

%In this work we applied CNNs to the problem of unconstrained ear recognition. The main problem that we observed is the lack of data, specifically for the recognition task in general it is difficult to obtain large scale datasets. Compared to some other datasets such as Flower dataset which has similar number of classes (102) of flowers there is 40--250 images per class, which is much more, especially if we consider that the inter-class variance between flowers is much greater (color, shape, etc.) compared to the ears that share quite a lot of similarities between persons. The next thing that we could do to improve performance would be to use different pre-learned models. Instead of using ImageNet model, we could learn the model on a binary classification task of ears and background to really fine-tune the features for ears and then fine-tune it on a recognition task. Exploiting different machine learning methods on features extracted by CNNs is also the possibility. In general, despite the lack of data, we got the results that are better than those of the best non-CNN based approaches. 
%Further analysis needs to be done, especially testing all approaches under the same methodology to get the proper comparison. ???

%In this work we applied CNNs to the problem of unconstrained ear recognition. The main problem that we observed is the lack of data, especially here for the ear recognition task where interclass differences are much smaller compared to other object recognition tasks, where images between different object vary significantly.

In this work we studied the problem of training CNN-based models for (closed set) ear recognition using limited training data. We investigated different strategies towards model training and were able to build a model that improved on the best performing state-of-the-art technique (based on HOG descriptors) included in our comparison by close to 30\% in terms of the Rank-1 recognition rate. 

The best model we were able to build was based on the SqueezeNet architecture. The model was initialize with parameters learned with the ImageNet data and then fine tuned using a limited set of 1383 ear images of 166 classes that were augmented by a factor of 100.  

In terms of future work, our goal will be to develop CNN-based models for open-set recognition problems and move beyond the close-set protocol explored in this paper. The difficulty with this approach is that the softmax layer at the top of the networks will have to be removed and image descriptors computed at the fully connected layers will have to be used as image representations. This task is typically more challenging than closed set recognition, as the computed descriptors need to be representative of unseen image classes (or identities), which commonly induces the need for even more training data.
%\addtolength{\textheight}{-3cm} 
%There is still room for improvement. One possibility would be to use different pre-learned models. Instead of using ImageNet model, we could learn the model on a binary classification task of ears and background to really fine-tune the features for ears and then fine-tune it on a recognition task. Furthermore, exploiting different architectures and merging them together, assembling larger and more robust pipeline would also be a possibility.

\bibliographystyle{IEEEtran}
\bibliography{bibliography}

\end{document}